\documentclass[runningheads]{llncs}
\usepackage[T1]{fontenc}
\usepackage{graphicx}
\usepackage{subfig}
\usepackage{comment}
\usepackage{booktabs}
\usepackage{adjustbox}

\usepackage{rotating} % <-- HERE

\usepackage[misc]{ifsym}

\usepackage{makecell}
\usepackage{array}  
\usepackage[table]{xcolor} % für Farben in Tabellen

\definecolor{lightgray}{gray}{0.95}

\begin{document}
\title{The Sound of Death:\\Deep Learning Reveals Vascular Damage from Carotid Ultrasound}
\titlerunning{Predicting All-Cause Death from Vascular Damage}

\author{Christoph Balada\inst{1,2}\thanks{\Letter~Corresponding author: 
\email{christoph.balada@dfki.de}}\orcidID{0000-0003-0307-7866} \and
Aida Romano-Martinez\inst{3,4} \and
Payal Varshney\inst{1,2} \and
Vincent ten Cate\inst{3,4} \and
Katharina Geschke\inst{3} \and
Jonas Tesarz\inst{3} \and
Paul Claßen\inst{3} \and
Alexander K. Schuster\inst{3} \and
%Karl Lackner\inst{2,3} \and
Dativa Tibyampansha\inst{3} \and
%Stavros Konstantinides\inst{2,3,4} \and
Karl-Patrik Kresoja\inst{3} \and
Philipp S. Wild\inst{3,4} \and
Sheraz Ahmed\inst{1} \and
Andreas Dengel\inst{1,2}\orcidID{0000-0002-6100-8255}}

%\authorrunning{Anonymous authors et al.}
\authorrunning{C. Balada et al.}

\institute{
German Research Center for Artificial Intelligence (DFKI), 67663 Kaiserslautern, Germany \and
RPTU University Kaiserslautern-Landau, 67663 Kaiserslautern, Germany \and
University Medical Center of the Johannes Gutenberg-University Mainz, Germany \and
German Center for Cardiovascular Research (DZHK), Germany}

\maketitle              % typeset the header of the contribution

\begin{abstract}
Cardiovascular diseases (CVDs) remain the leading cause of mortality worldwide, yet early risk detection is often limited by available diagnostics.
Carotid ultrasound, a non-invasive and widely accessible modality, encodes rich structural and hemodynamic information that is largely untapped. 
Here, we present a machine learning (ML) framework that extracts clinically meaningful representations of vascular damage (VD) from carotid ultrasound videos, using hypertension as a weak proxy label.
The model learns robust features that are biologically plausible, interpretable, and strongly associated with established cardiovascular risk factors, comorbidities, and laboratory measures. 
High VD stratifies individuals for myocardial infarction, cardiac death, and all-cause mortality, matching or outperforming conventional risk models such as SCORE2. 
Explainable AI analyses reveal that the model relies on vessel morphology and perivascular tissue characteristics, uncovering novel functional and anatomical signatures of vascular damage.
This work demonstrates that routine carotid ultrasound contains far more prognostic information than previously recognized. Our approach provides a scalable, non-invasive, and cost-effective tool for population-wide cardiovascular risk assessment, enabling earlier and more personalized prevention strategies without reliance on laboratory tests or complex clinical inputs.

\end{abstract}

%%%%%%%%%%%%%%%%%%%%%%%%%%%%%%%%%%%%%%%%%%%%%%%%%%%%%%%%%%%%%%%%%%%%%%%%%%%%%%%%%%%%%%%%%%%%%%%%%%%
%% Introduction
%%%%%%%%%%%%%%%%%%%%%%%%%%%%%%%%%%%%%%%%%%%%%%%%%%%%%%%%%%%%%%%%%%%%%%%%%%%%%%%%%%%%%%%%%%%%%%%%%%%
\section{Introduction}
Cardiovascular diseases (CVDs) remain the leading cause of mortality worldwide, accounting for approximately 17.9 million deaths in 2019, which corresponds to 32\% of all global deaths \cite{CVDfactSheet}. 
The majority of CVDs are preventable through the mitigation of behavioural and environmental risk factors, including tobacco use, unhealthy diet, obesity, physical inactivity, harmful alcohol consumption, and exposure to air pollution \cite{rajagopalan2018air}. 
Early detection of CVD is therefore critical, as timely intervention through lifestyle counselling and pharmacological treatment can substantially improve patient outcomes.

In recent years, Machine Learning (ML) has emerged as a promising approach for enabling the early and individualized detection of CVD risk, offering the potential to uncover subtle patterns in physiological data that may elude traditional clinical assessment. 
Concurrently, the implementation of ML has the potential to facilitate the optimisation of evaluation processes and enable the execution of screening initiatives on a broader scale.
Furthermore, ML can mitigate limitations of clinic-based measurements, such as the white-coat effect, where elevated blood pressure readings occur in clinical settings but not in everyday environments \cite{franklin2013white}.

However, the application of ML in medicine is often constrained by limited data availability and imperfect labels. 
Clinical datasets are costly to acquire, labor-intensive to annotate, and frequently affected by diagnostic uncertainty or proxy-based labels that only indirectly reflect the underlying pathology. 
These limitations make it challenging to capture subtle disease processes unless models can exploit richer, underutilized information sources.

Against this backdrop, carotid ultrasound emerges as an especially promising modality for ML-based CVD risk assessment. 
Contemporary clinical practice focuses primarily on intima-media thickness, arterial stiffness, and the presence of plaques or stenosis \cite{bao2023carotid}. 
Yet, carotid ultrasound—particularly in video format—contains far richer structural and temporal information that may encode subtle vascular abnormalities not captured by conventional metrics. 
These include dynamic hemodynamic disturbances \cite{abbott2015systematic}, transient thrombotic phenomena such as spontaneous “fogging” \cite{black2000spontaneous}, and tissue-level alterations in the perivascular region \cite{martinez2018hypertension,zhang2021association,baradaran2018association}. 
Given this context, ML offers a unique opportunity to uncover latent features embedded in carotid ultrasound videos that could substantially enhance cardiovascular risk assessment.

In this study, we investigate the capacity of ML to uncover clinically meaningful vascular features from carotid ultrasound videos, aiming to extend risk stratification beyond the constraints of traditional approaches. 
Using data from the Gutenberg Health Study \cite{wild2012gutenberg}, we train a neural network on hypertension as a weak proxy label for underlying vascular pathology. 
Despite the inherent noise and limited specificity of this target, the model learns representations that are biologically plausible, visually interpretable, and strongly associated with established cardiovascular biomarkers and risk factors.

These learned features show robust correlations with major adverse cardiovascular events and enable effective stratification of individuals into low- and high-risk groups for stroke, myocardial infarction (MI), cardiac death, and all-cause mortality. 
Notably, the model performs on par with—or surpasses—conventional risk prediction algorithms such as SCORE2 \cite{esc2021score2}, despite relying solely on carotid ultrasound imaging without laboratory measurements or additional clinical data.

Because absolute blood pressure cannot be inferred from grayscale B-mode ultrasound in the absence of Doppler information, we hypothesize that the model detects structural and functional vascular abnormalities linked to hypertension rather than the blood pressure value itself. Accordingly, individuals classified as hypertensive by the model are interpreted as exhibiting elevated vascular damage (VD), a representation that aligns with the observed prognostic patterns.

Crucially, this level of predictive performance is achieved using only a single, widely available, non-invasive, and low-cost imaging modality—carotid sonography—highlighting its untapped potential as a scalable tool for population-level cardiovascular risk assessment.
In addition, the application of Explainable Artificial Intelligence (XAI) methods, particularly occlusion-based attribution and counterfactual examples, reveals that the model employs vessel morphology and surrounding tissue structures for prediction.
This enhancement not only improves the interpretability of the system but also identifies novel anatomical and functional signatures of VD. 
Ultimately, it is demonstrated that conventional carotid sonography assessments do not fully utilise the potential of features captured in the ultrasound modality.

%%%%%%%%%%%%%%%%%%%%%%%%%%%%%%%%%%%%%%%%%%%%%%%%%%%%%%%%%%%%%%%%%%%%%%%%%%%%%%%%%%%%%%%%%%%%%%%%%%%
%% Results
%%%%%%%%%%%%%%%%%%%%%%%%%%%%%%%%%%%%%%%%%%%%%%%%%%%%%%%%%%%%%%%%%%%%%%%%%%%%%%%%%%%%%%%%%%%%%%%%%%%
\section{Results}
\subsection{Patient characteristics}
We analysed longitudinal carotid ultrasound video data from the initial, 5-year and 10-year follow-up assessments of the Gutenberg Health Study. 
At the initial assessment, of the 14,246 participants, 7,135 (50.1\%) were classified as hypertensive based on measured and self-reported diagnosis.
At 5-year follow-up, among the 9,815 individuals who returned, 5,302 (54.0\%) were classified as hypertensive and at 10-year follow-up, among the 9,115 individuals who returned, 5,218 (57.2\%) were classified as hypertensive.
A comprehensive statistical summary of the participants' clinical characteristics across all assessments is provided in Table \ref{tab:participants}.

\begin{sidewaystable} % <-- HERE
%\begin{table} % <-- HERE

\caption{
{\large Dataset Characteristics} \\
{\small Gutenberg Health Study - Participants with Carotid Ultrasound}
} 
\label{tab:participants}

\fontsize{5.0pt}{6.0pt}\selectfont
\rowcolors{5}{lightgray}{white} % ab Zeile 3 (damit Kopf unberührt bleibt)

\begin{tabular*}{\linewidth}{@{\extracolsep{\fill}}p{3cm}rrrrrrrrr}
\toprule
 & \multicolumn{3}{c}{Initial Assessment} & \multicolumn{3}{c}{5-years Follow-Up} & \multicolumn{3}{c}{10-years Follow-Up} \\ 
\cmidrule(lr){2-4} \cmidrule(lr){5-7}
\makecell{Characteristics} 

& \makecell{All \\ N=14246} 
& \makecell{Hypertensive \\ N=7135} 
& \makecell{Non-\\Hypertensive \\ N=7111} 

& \makecell{All \\ N=9815} 
& \makecell{Hypertensive \\ N=5302} 
& \makecell{Non-\\Hypertensive \\ N=4513}

& \makecell{All \\ N=9115} 
& \makecell{Hypertensive \\ N=5218} 
& \makecell{Non-\\Hypertensive \\ N=3897} \\ 

\midrule\addlinespace[2.5pt]

\makecell{Age} & & & & & & & & & \\
35-45 & 3100 (21.8\%) & 637 (8.9\%) & 2459 (34.6\%) & 1002 (10.2\%) & 205 (3.9\%) & 795 (17.6\%) & 0 (0.0\%) & 0 (0.0\%) & 0 (0.0\%) \\
45-55 & 3751 (26.3\%) & 1523 (21.3\%) & 2225 (31.3\%) & 2616 (26.7\%) & 928 (17.5\%) & 1687 (37.4\%) & 2259 (24.8\%) & 790 (15.1\%) & 1465 (37.7\%) \\
55-65 & 3791 (26.6\%) & 2320 (32.5\%) & 1470 (20.7\%) & 2699 (27.5\%) & 1548 (29.2\%) & 1150 (25.5\%) & 2767 (30.4\%) & 1427 (27.3\%) & 1335 (34.4\%) \\
65-75 & 3604 (25.3\%) & 2655 (37.2\%) & 949 (13.4\%) & 2552 (26.0\%) & 1862 (35.1\%) & 689 (15.3\%) & 2506 (27.5\%) & 1746 (33.5\%) & 758 (19.5\%) \\
75-85 & 0 (0.0\%) & 0 (0.0\%) & 0 (0.0\%) & 946 (9.6\%) & 759 (14.3\%) & 186 (4.1\%) & 1562 (17.1\%) & 1239 (23.7\%) & 322 (8.3\%) \\

\makecell{Sex} & & & & & & & & & \\
Men & 7199 (50.5\%) & 3953 (55.4\%) & 3244 (45.7\%) & 5014 (51.1\%) & 2947 (55.6\%) & 2063 (45.8\%) & 4705 (51.6\%) & 2961 (56.7\%) & 1739 (44.8\%) \\
Women & 7047 (49.5\%) & 3182 (44.6\%) & 3859 (54.3\%) & 4801 (48.9\%) & 2355 (44.4\%) & 2444 (54.2\%) & 4410 (48.4\%) & 2257 (43.3\%) & 2146 (55.2\%) \\

\makecell{Comorbidities} & & & & & & & & & \\
CVD & 1695 (11.9\%) & 1223 (17.1\%) & 472 (6.6\%) & 805 (8.2\%) & 649 (12.2\%) & 156 (3.5\%) & 926 (10.2\%) & 719 (13.8\%) & 205 (5.3\%) \\
CAD & 617 (4.3\%) & 490 (6.9\%) & 127 (1.8\%) & 245 (2.5\%) & 211 (4.0\%) & 34 (0.8\%) & 228 (2.5\%) & 198 (3.8\%) & 30 (0.8\%) \\
CHF & 191 (1.3\%) & 150 (2.1\%) & 41 (0.6\%) & 145 (1.5\%) & 121 (2.3\%) & 24 (0.5\%) & 193 (2.1\%) & 160 (3.1\%) & 33 (0.8\%) \\
AFIB & 388 (2.7\%) & 277 (3.9\%) & 111 (1.6\%) & 261 (2.7\%) & 207 (3.9\%) & 54 (1.2\%) & 282 (3.1\%) & 218 (4.2\%) & 62 (1.6\%) \\
History of stroke & 268 (1.9\%) & 203 (2.8\%) & 65 (0.9\%) & 93 (0.9\%) & 77 (1.5\%) & 16 (0.4\%) & 117 (1.3\%) & 97 (1.9\%) & 20 (0.5\%) \\
History of myocardial infarction & 424 (3.0\%) & 309 (4.3\%) & 115 (1.6\%) & 108 (1.1\%) & 93 (1.8\%) & 15 (0.3\%) & 103 (1.1\%) & 88 (1.7\%) & 15 (0.4\%) \\

\makecell{Traditional \\ Risk Markers} & & & & & & & & & \\
Hypertension & 7135 (50.1\%) & 7135 (100.0\%) & 0 (0.0\%) & 5302 (54.0\%) & 5302 (100.0\%) & 0 (0.0\%) & 5218 (57.2\%) & 5218 (100.0\%) & 0 (0.0\%) \\
Current Smoking & 2765 (19.4\%) & 1095 (15.3\%) & 1667 (23.5\%) & 1504 (15.3\%) & 674 (12.7\%) & 830 (18.4\%) & 1127 (12.4\%) & 545 (10.4\%) & 579 (14.9\%) \\
Diabetes Mellitus type 2 & 1340 (9.4\%) & 1062 (14.9\%) & 278 (3.9\%) & 1027 (10.5\%) & 859 (16.2\%) & 168 (3.7\%) & 1023 (11.2\%) & 849 (16.3\%) & 174 (4.5\%) \\
Dyslipidemia & 4938 (34.7\%) & 3123 (43.8\%) & 1813 (25.5\%) & 3357 (34.2\%) & 2311 (43.6\%) & 1044 (23.2\%) & 3007 (33.0\%) & 2182 (41.8\%) & 824 (21.2\%) \\
\bottomrule
\end{tabular*}
\end{sidewaystable} % <-- HERE
%\end{table}

\subsection{Model Performance}
\begin{figure}[!t]
 \centering
    \includegraphics[width=1\linewidth,trim=0cm 0cm 0cm 0cm, clip]{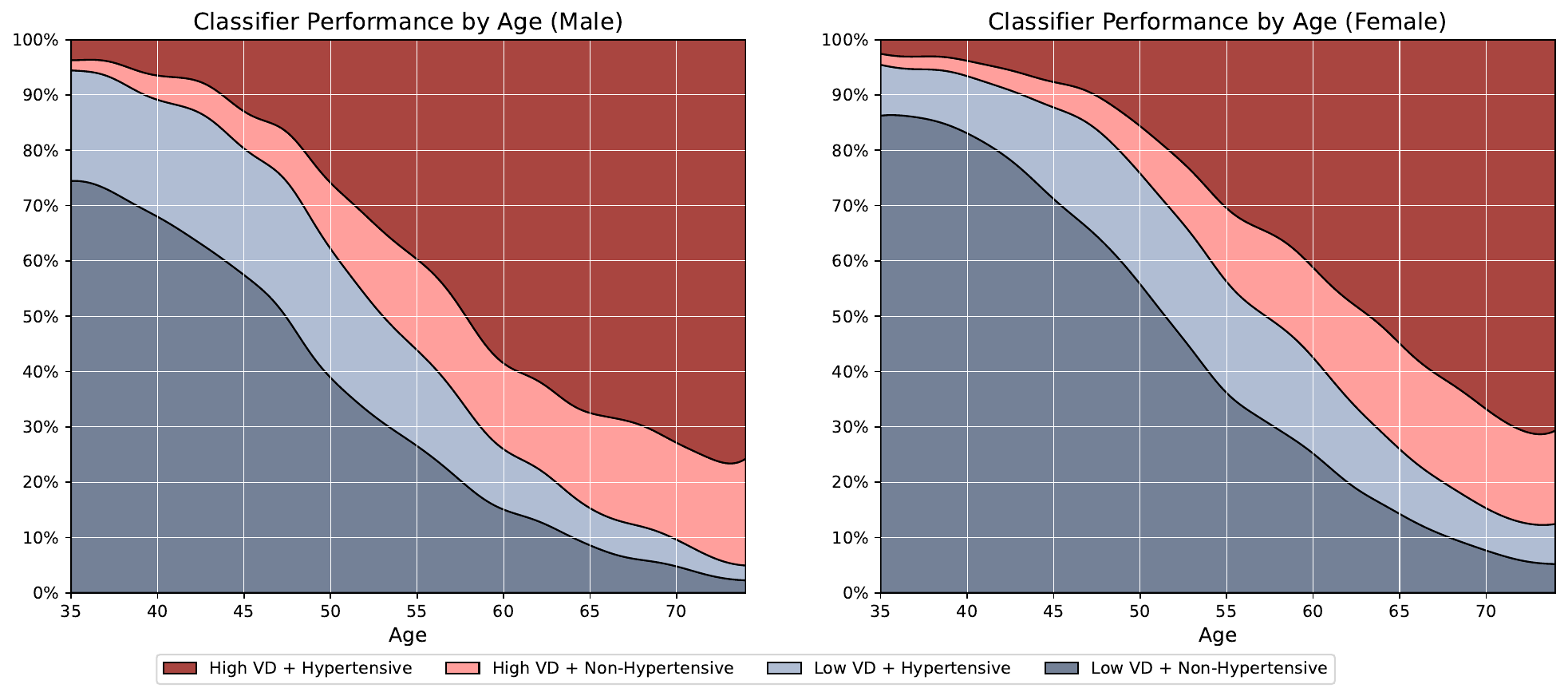}
    
    \caption{Classification accuracy by age at initial assessment.
    Left: male participants; right: female participants. Dark colours denote correct classifications, with true positives (high vascular damage (VD) and hypertensive) shown in red and true negatives (low VD and non-hypertensive) in blue. Light shades indicate misclassifications, specifically false positives (light red; high VD and non-hypertensive) and false negatives (light blue; low VD and non-hypertensive). Overall, classification patterns between males and females are comparable above 55 years of age. However, among younger individuals (<45 years), males exhibit a higher proportion of false negatives, accompanied by a reduced rate of true negatives compared with females.
}%
    \label{fig:kde_performance}%
\end{figure}

Using VideoMAE\cite{tong2022videomae}, we achieved a balanced accuracy of 72.3\% when aggregating predictions across multiple clips per video and across multiple videos per individual, using simple averaging of model confidence scores. 
Figure \ref{fig:kde_performance} depicts performance trends stratified by age and reveals several clinically relevant patterns, particularly in male participants.

Notably, among younger individuals (below 45 years), males exhibit a higher proportion of low VD with hypertension, accompanied by a reduced rate of low VD without hypertension compared with females.
Furthermore, a more pronounced decline in the prevalence of low VD is observed among males, resulting in a discrepancy of approximately 16\% between males (26\% low VD) and females (42\% low VD) at the age of 60.
No major performance differences were observed between initial and follow-up assessments aside from the expected age progression.

\subsection{Feature Interpretation}
To clarify the decision-making process of our model and assess whether its predictions rely on physiologically meaningful image content, we employed two complementary explainable AI techniques: occlusion-based feature attribution and counterfactual example generation. 
Both analyses consistently identified the vessel wall and adjacent perivascular tissue as the most influential regions for model predictions.

\begin{figure}[!t]
 \centering
    \includegraphics[width=1\linewidth,trim=0.0cm 18cm 0cm 0cm,clip]{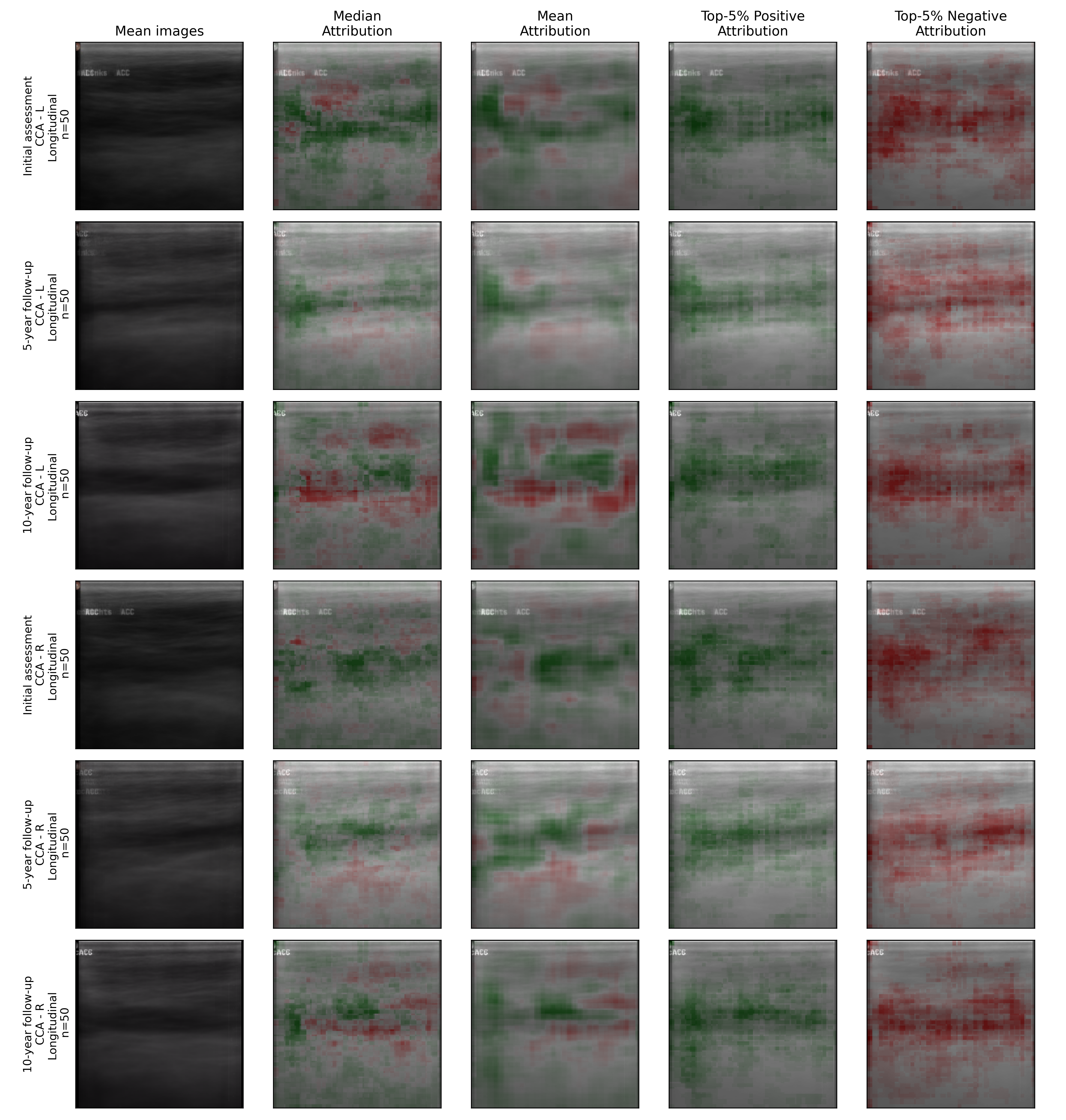}
    
    \caption{Average, median, top-5\% positive and negative regions of interest (ROIs) of 50 videos. ROIs highlight which regions have been found most important for the model to predict an example as high vascular damage (green) or as low vascular damage (red). Attributions and background images have been averaged over 50 videos, selected around the global dataset mean. Progress in course of the different assessments is illustrated per row. In all cases the vessel walls are rendered most important. However, in particular in older ages (e.g. 10-year follow-up), ROIs are in particular pronounced in the perivascular tissue. 
    }%
    \label{fig:occ_attrib}%
\end{figure}

Occlusion-based feature attribution, averaged across 50 participants for each of the three assessment time points, revealed a robust focus on the vessel wall (Fig. \ref{fig:occ_attrib}). 
Notably, we observed an assessment-dependent shift in the distribution of salient regions: whereas the baseline and 5-year follow-up assessments showed dominant attention on the vessel wall, the 10-year follow-up demonstrated a pronounced expansion of salient regions into the perivascular tissue. 
This shift suggests that, in older participants, the model increasingly leverages textural cues in the surrounding tissue like perivascular fat.

Both explainability approaches further revealed that the spatial configuration and intensity of the identified regions of interest varied over the course of the ultrasound sequences, reflecting dynamic changes during the cardiac cycle. 
It was also investigated whether model performance or the regions of interest were affected if the on-screen ECG trace was retained during preprocessing. 
Ultimately, the integration of the ECG trace did not yield substantial benefits when compared to the model trained without it.

\subsection{Analysis of Clinical Data} % Clinical Analysis? Clinical Test? Clinical Results?
To evaluate the clinical utility of VD derived from the model’s prediction, we performed a multi-faceted analysis, including (i) statistical comparison across clinical parameters and biomarkers, (ii) survival analysis using Kaplan–Meier estimators, and (iii) risk modelling via Cox proportional hazards models.

\begin{figure*}[!t]
 \centering
    \includegraphics[width=1.0\linewidth,trim=0cm 0cm 0cm 0cm]{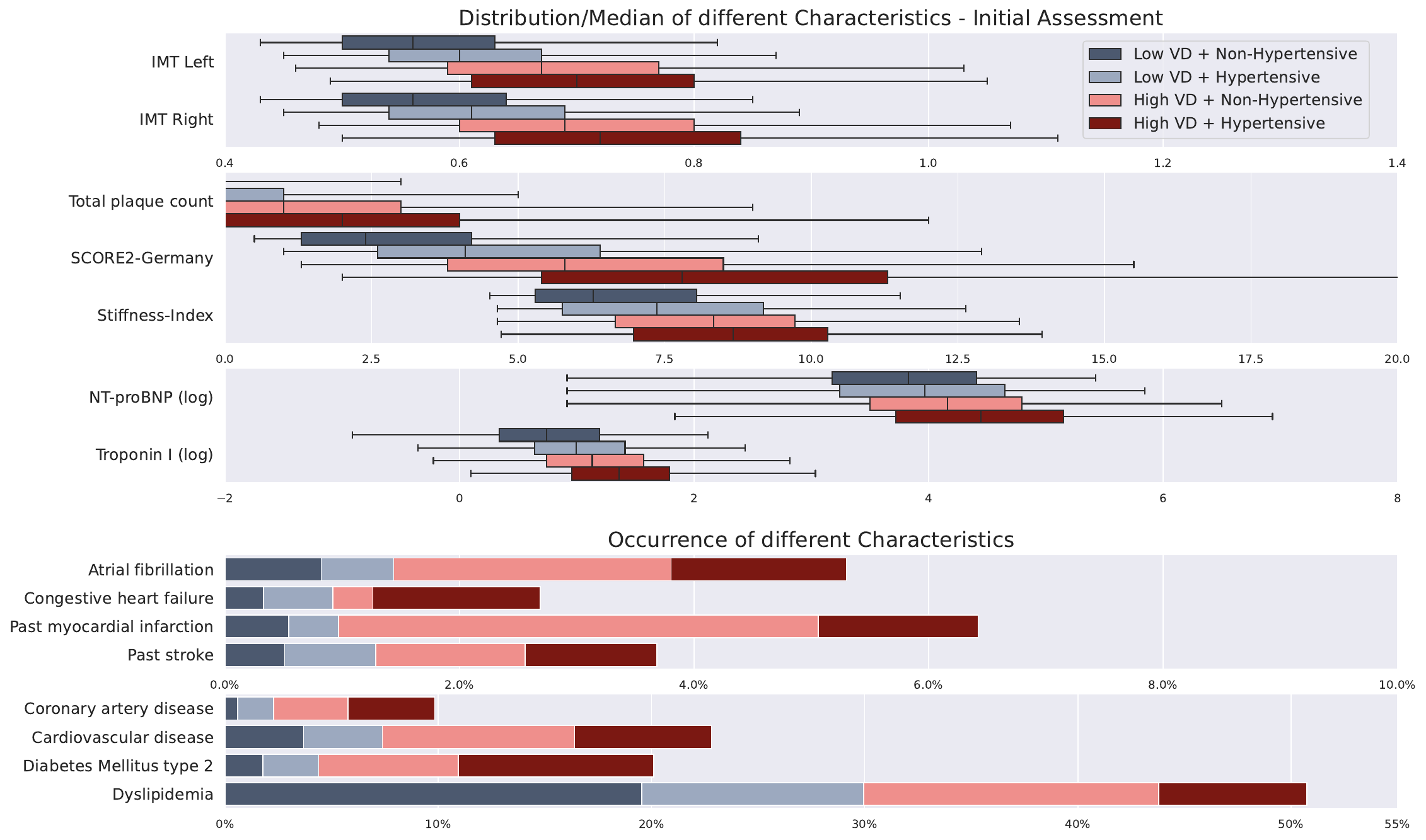} %
    \qquad 
    \caption{Comparison of clinical parameters in the Gutenberg Health Study cohort.
    Higher values generally indicate poorer cardiovascular status. 
    Individuals with high vascular damage, irrespective of hypertension, consistently exhibit the poorest overall condition.
    }%
    \label{fig:param_comp}%
\end{figure*}

\subsubsection{Clinical Parameter Association}
We stratified participants into four groups based on their predicted VD (high vs. low) and hypertensive versus non-hypertensive.
Figure \ref{fig:param_comp} compares these groups across a broad spectrum of cardiovascular health parameters. 
Across all markers, individuals with high VD and hypertension exhibited the poorest health status. 
Interestingly, individuals with high VD but who were not hypertensive consistently displayed the second-worst profile. 
This gradient was consistently reflected in all individual parameters and composite risk scores such as SCORE2.

The following findings merit particular attention: In contrast to individuals with low VD and no hypertension, those with high VD and hypertension exhibited around an 8-fold higher prevalence of prior MI$[8.5,11.0]$ and heart failure$[5.4,8.3]$.
Moreover, our VD marker also stratifies participants in terms of IMT and arterial stiffness, both of which are traditionally used as markers for increased CVD risk.

\begin{figure*}[!b]
 \centering
    \includegraphics[width=1.0\linewidth]{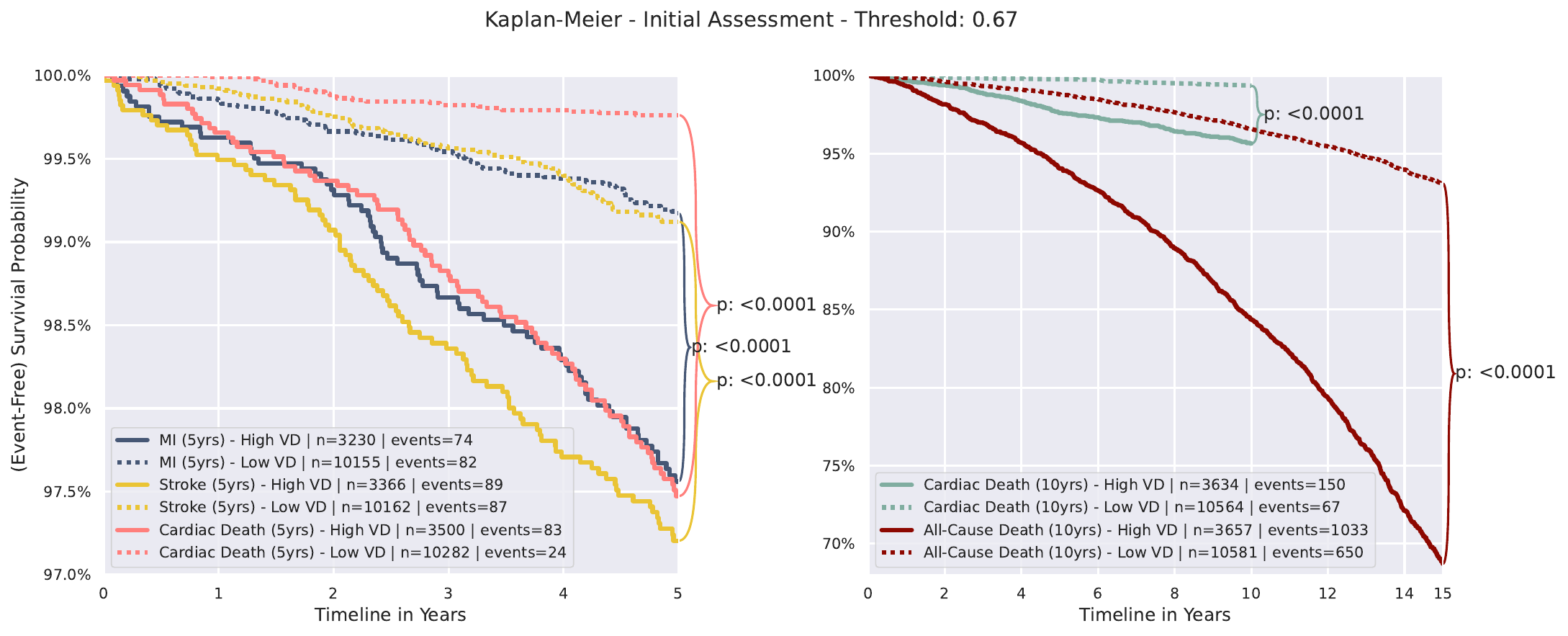} %
    \qquad
    \includegraphics[width=1.0\linewidth]{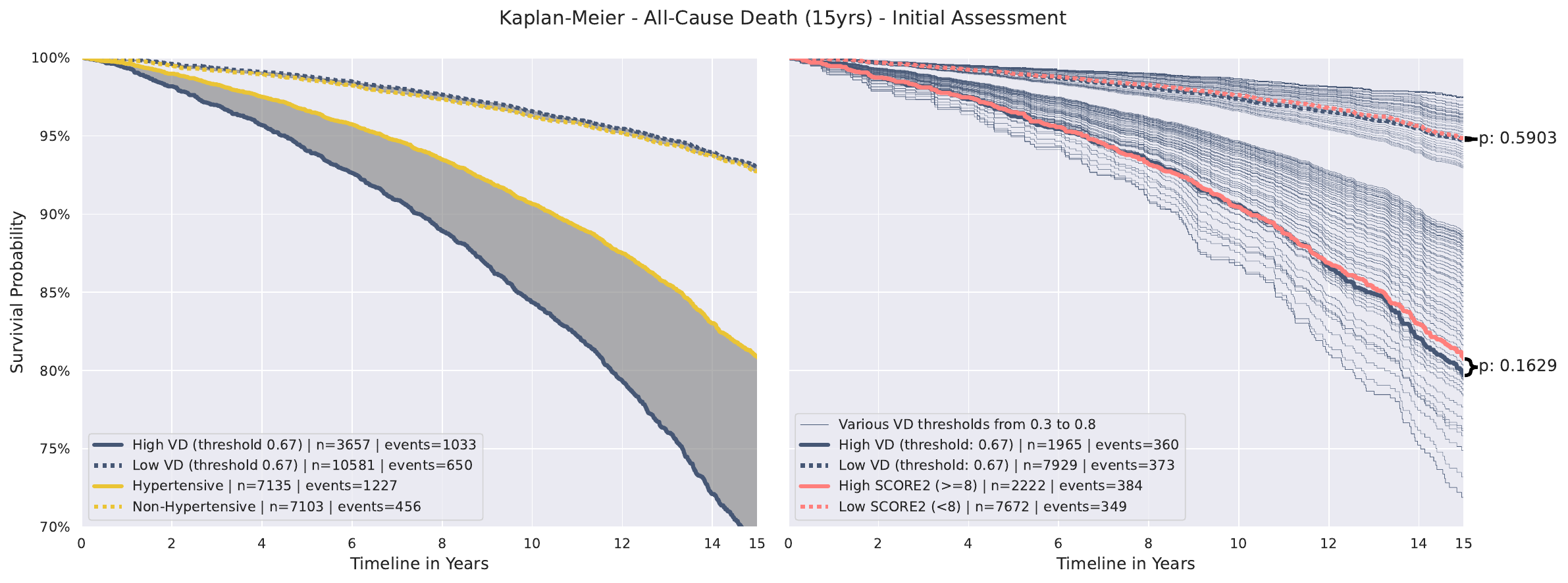} %
    
    \caption{Kaplan–Meier estimates for vascular damage (VD).
    Top: (Event-free) survival for incident myocardial infarction, stroke, and cardiac death over a 5-year period (left) and for cardiac and all-cause death over 10 years (right), comparing participants with high VD (solid lines) versus low VD (dashed lines).    
    Bottom: 15-year all-cause death stratified by VD and hypertension label used during model training (left) and by VD versus SCORE2 baseline risk model (right). VD consistently outperformed hypertension status. 
    High VD at the predefined threshold of $0.67$ exhibited risk-stratification characteristics, closely comparable to a high SCORE2.
    }
    \label{fig:kaplanmeier_combi}%
\end{figure*}

\begin{figure*}[!t]
 \centering
    \subfloat[\centering Model Performance (All-Cause-Death 15yrs)]{{\includegraphics[width=1.\linewidth,trim=0cm 0cm 0cm 0cm,clip]{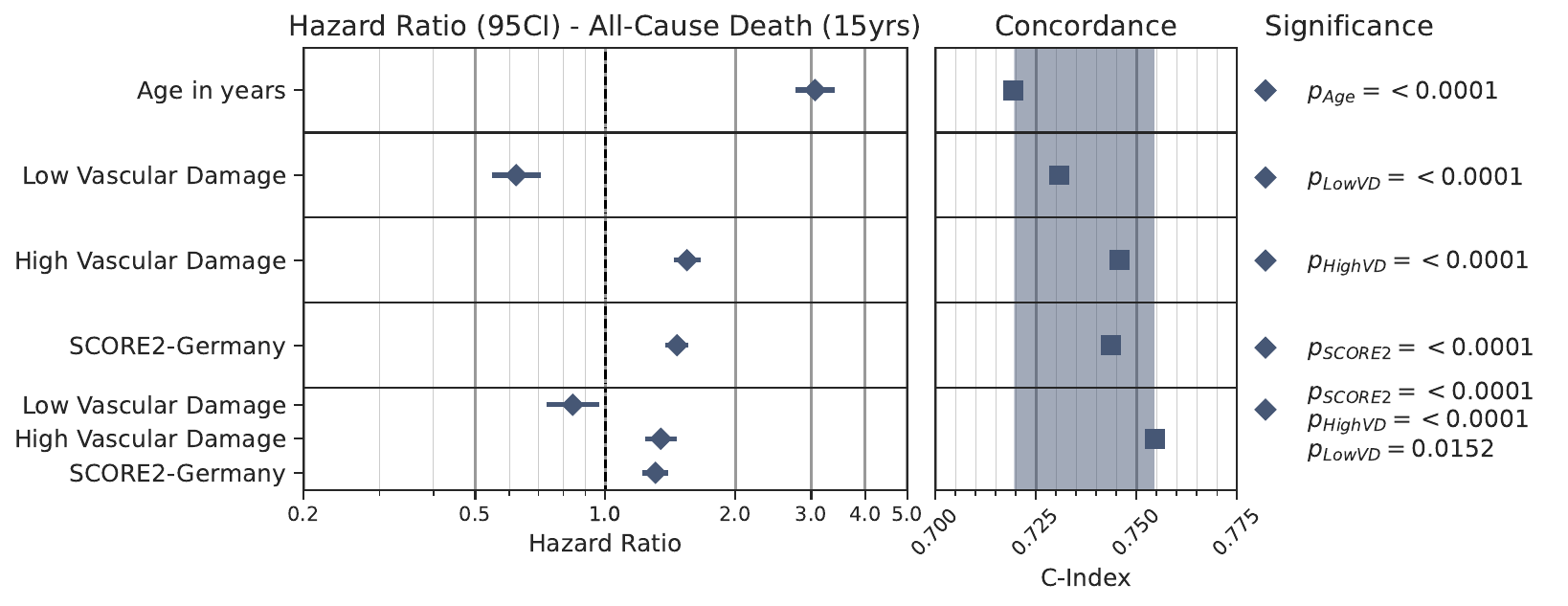} }}%
    \qquad
    \subfloat[\centering C-index Comparison]{{\includegraphics[width=.8\linewidth,trim=0cm 0cm 0cm 0cm,clip]{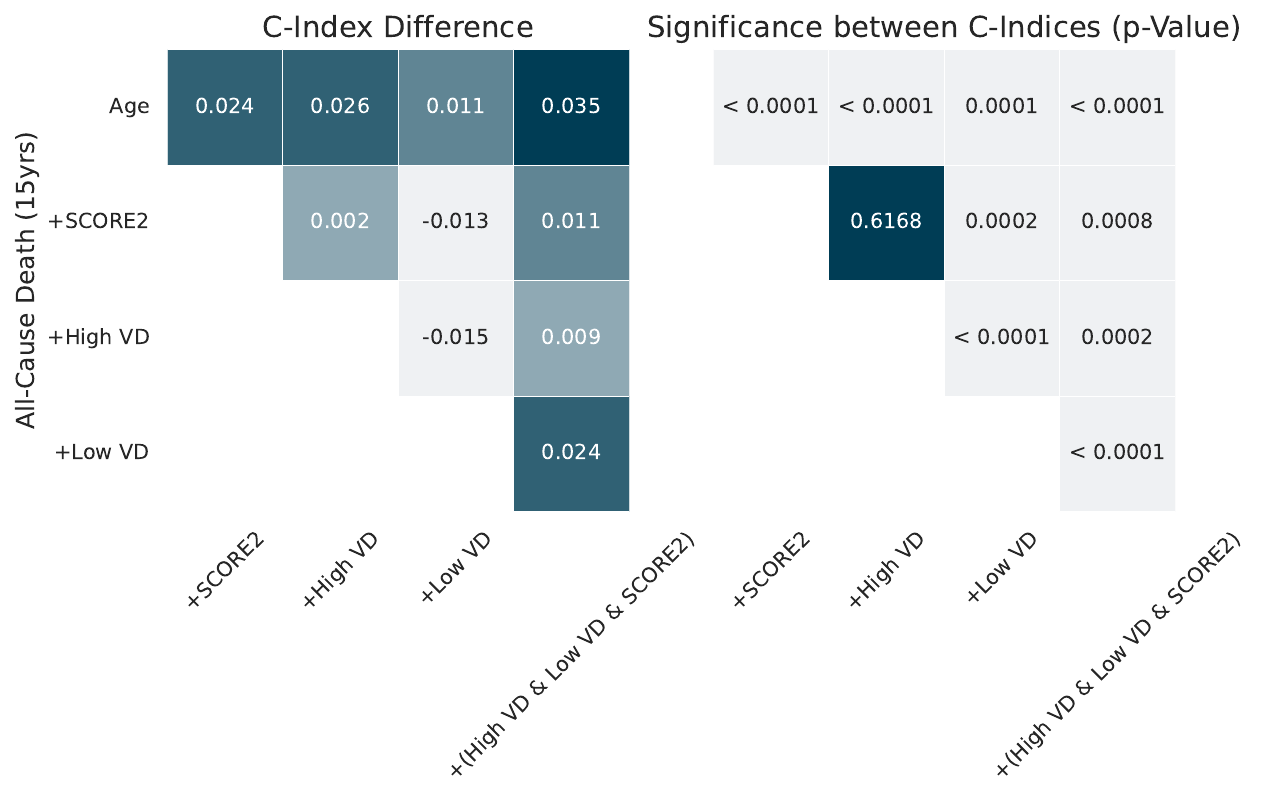} }}%

    \caption{
    Cox proportional hazards models incorporating different covariates. 
    (a) Hazard ratios (left) and corresponding concordance and significance (right)
    The shaded regions highlight the difference between the lowest and highest C-index.
    Combining high and low vascular damage score and SCORE2, shows the best performance. 
    (b) Differences in concordance index (C-index) (left) and statistical significance (right). 
    }%
    \label{fig:cox_score2}%
\end{figure*}

\begin{figure}[!t]
 \centering
    \includegraphics[width=1.\linewidth,trim=0cm 0cm 0cm 0cm,clip]{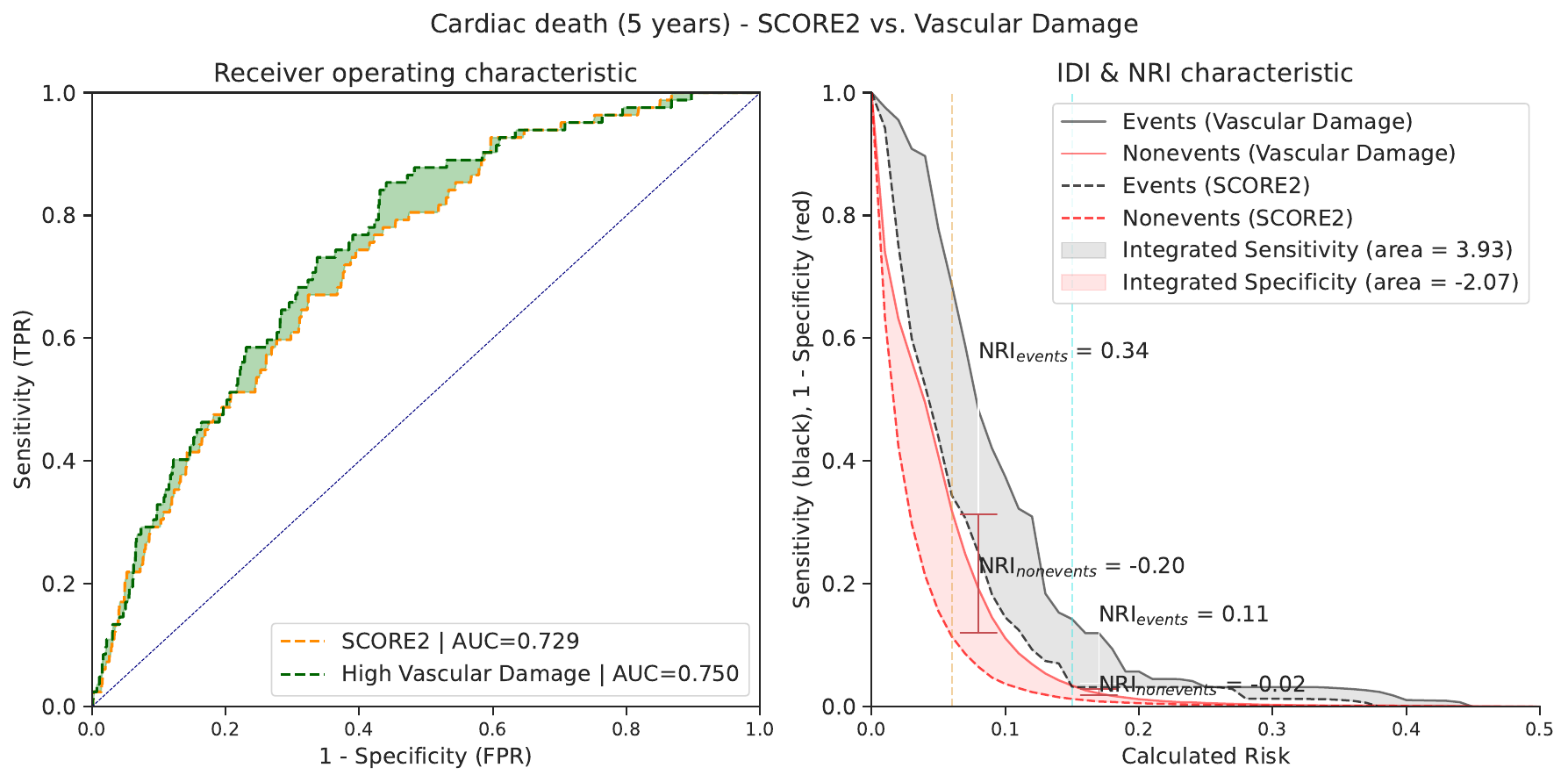}
    
    \caption{
    Left: Receiver operating characteristic (ROC) curves for a linear classifier predicting 5-year cardiac death based on the predicted hazard from the vascular damage (VD) and SCORE2 Cox models. Right: Integrated Discrimination Improvement (IDI) and Net Reclassification Improvement (NRI) comparing both models. The VD-based model demonstrated superior discrimination, achieving higher area under the ROC curve, as well as improved IDI and NRI, indicating enhanced reclassification performance for the fatal 5-year cardiac death subgroup.
    }%
    \label{fig:nri_score2}%
\end{figure}

\subsubsection{Survival Analysis}
To assess prognostic relevance, we applied Kaplan–Meier estimators to model (event-free) survival across multiple endpoints. 
As shown in Figure \ref{fig:kaplanmeier_combi}, our VD marker robustly stratified risk for major events including incident myocardial infarction, stroke, and cardiac death over 5 years, as well as for cardiac and all-cause death over 10 and 15 years.

Furthermore, we compared VD-based stratification with traditional indicators, including hypertension and elevated SCORE2 risk, in relation to 15-year all-cause mortality. 
Across all analyses, our VD marker outperformed hypertension as a discriminator of long-term risk. 
Notably, hypertension was the label used to train the model, making the superior performance of VD particularly remarkable: despite being derived from the same data, VD provides a more accurate measure of long-term risk than the training label itself. 
A high VD value at the predefined threshold of $0.67$ exhibited risk-stratification characteristics closely comparable to those of a high SCORE2 category—yet without the need for laboratory measurements or the applicability constraints inherent to SCORE2.
Importantly, combining VD with SCORE2 provided the most robust separation between high- and low-risk groups, indicating that VD offers complementary prognostic information beyond conventional clinical risk scores.

\subsubsection{Hazard Modelling}
VD markedly outperforms traditional risk indicators in long-term mortality prediction. 
Using Cox proportional hazards models, we leveraged the predicted confidence for high and low VD rather than binary classification. Incorporation of these confidence scores improved the concordance index (C-index) by up to $0.11$ over an age-only baseline.
Combining VD with SCORE2 further enhanced predictive accuracy across nearly all endpoints, achieving a $0.035$ C-index gain for 15-year all-cause mortality (Figure \ref{fig:cox_score2}).
High and low VD scores were strongly associated with 15-year all-cause death, 5- and 10-year cardiac death, and 5-year myocardial infarction, while low VD also predicted 5-year stroke. 
High VD alone improved the C-index for 5-year cardiac death by $0.048$ over age and $0.026$ over SCORE2.

Using predicted hazards from the Cox models as input for a linear classifier of 5-year cardiac death, the VD-based model achieved a higher area under the ROC curve ($0.75$) than the SCORE2-based model ($0.729$), primarily through enhanced sensitivity in identifying high-risk individuals.
Net reclassification improvement (NRI) analyses confirmed substantial gains in risk stratification, up to $0.34$, underscoring the added prognostic value of VD beyond conventional clinical risk scores (Figure \ref{fig:nri_score2}).

%%%%%%%%%%%%%%%%%%%%%%%%%%%%%%%%%%%%%%%%%%%%%%%%%%%%%%%%%%%%%%%%%%%%%%%%%%%%%%%%%%%%%%%%%%%%%%%%%%%
%% Discussion
%%%%%%%%%%%%%%%%%%%%%%%%%%%%%%%%%%%%%%%%%%%%%%%%%%%%%%%%%%%%%%%%%%%%%%%%%%%%%%%%%%%%%%%%%%%%%%%%%%%
\section{Discussion}
Despite using noisy hypertension labels as a proxy task, our model demonstrated solid classification performance, achieving a balanced accuracy of 72.3\% on the combined test set when aggregating predictions across available clips. 
However, the primary objective of this study is not accurate hypertension detection, but rather the extraction of features relevant to vascular and cardiovascular health using easily accessible proxy labels. 
In this context, cases where the model’s predictions diverge from the assigned labels—such as detecting high VD in non-hypertensive individuals—become particularly informative. 
As illustrated in Figure~\ref{fig:kde_performance}, the model’s behaviour aligns with the expected age-related development of vascular pathology. 
Young individuals with hypertension often lack observable VD, while older individuals frequently show VD regardless of hypertensive status, reflecting well-established progression patterns of arterial ageing \cite{cushman2003burden,jani2006ageing,mcgrath1998age}.
In light of the findings, we hypothesise that the model captures subclinical or unrecognised VD such as is typically the result of hypertension, emphasising the potential of carotid ultrasound to extend beyond its current applications.

Interestingly, the classification behaviour exhibited sex-specific patterns. 
While model performance among individuals aged >65 years was largely comparable between sexes, younger females demonstrated a lower prevalence of both hypertension and high VD compared with males. 
However, these differences gradually diminished with advancing age. 
This observation is consistent with the notion of hormonal protection against early atherosclerotic changes in premenopausal women \cite{davezac2021estrogen}.
The earlier onset and higher prevalence of hypertension in men likely contribute to the earlier manifestation of VD in this group. 
The convergence of patterns at older ages may reflect hormonal changes associated with menopause, as well as the increased initiation of antihypertensive therapy in older adults in Germany \cite{van2013prevalence}.

Feature analysis by XAI methodology revealed that the model does not rely solely on the carotid vessel or its wall, but also incorporates surrounding tissue when forming predictions. 
This was evident from both occlusion-based and counterfactual methods. 
While changes to the vessel wall are intuitive and also typically considered in conventional metrics like the carotid intima-media thickness, the surrounding perivascular tissue is not normally interpreted in carotid sonography, but could reflect pathological cellular changes as reported \cite{martinez2018hypertension,zhang2021association,baradaran2018association,vlodaver1968coronary}. 
In high VD cases, distinct spatially localized hotspots of relevance were observed throughout the cardiac cycle, often extending beyond the vessel itself.
This broader spatial focus offers a new perspective, as clinical assessments typically prioritize only the vessel and its immediate boundaries. 

The prediction confidence of the ML model varied dynamically throughout the ultrasound video and, consequently, across different phases of the cardiac cycle. 
This pattern suggests that periodic vascular compression and relaxation during systole and diastole could influence the ultrasound properties captured by the model.
These findings imply that the model may inherently exploit biomechanical information analogous to elastography, effectively leveraging cardiac-cycle–induced tissue deformation to infer vascular characteristics relevant to biological ageing.
The progression of this capability will be investigated in future research, with particular attention to temporal development of predictions across the ultrasound video, the cardiac cycle, and the integration and synchronization of the on-screen ECG trace.

The clinical relevance of the extracted VD marker was further confirmed through survival analyses and hazard modelling. 
Kaplan–Meier estimates consistently stratified individuals into low- and high-risk groups across multiple outcomes, including myocardial infarction, stroke, cardiac death, and all-cause mortality. 
Importantly, this stratification was not simply a surrogate for age: Cox proportional hazards models demonstrated that both high and low VD contributed significant predictive value even after adjusting for age.

By adjusting the classification threshold, high VD can be aligned with the risk stratification properties of SCORE2—without requiring laboratory measurements and without the age- and condition-specific limitations of SCORE2, which applies only to individuals aged 40–69 years without prior CVD or diabetes. 
Leveraging the model’s predicted confidence rather than binary classification further enhanced VD’s utility in hazard models. 
VD showed particularly strong and significant associations with 5-year cardiac death and 15-year all-cause mortality. 
Across all settings, VD consistently improved model performance, outperforming SCORE2 by up to $0.048$ in C-index for 5-year cardiac death and yielding stronger hazard ratios.

Crucially, VD does not rely on laboratory tests and is not constrained by age or CVD-specific eligibility criteria, making it a broadly applicable and cost-effective prognostic tool. 
Moreover, combining VD with SCORE2 produced the highest overall predictive accuracy, highlighting a synergistic benefit of integrating imaging-derived markers with conventional risk scores.

\section{Strengths and limitations}
This study leverages a large, deeply phenotyped cohort with an extensive collection of carotid ultrasound recordings to train a validated transformer-based architecture for the prediction of VD. 
In addition, state-of-the-art explainable AI techniques were employed to identify and visualize carotid features that differentiate individuals with high versus low VD, enhancing clinical interpretability. 
However, several limitations warrant consideration. 
The acquisition of deeply phenotyped cohorts with long-term follow-up is resource intensive, and therefore independent external validation of the model was not feasible within the current study. 
Nonetheless, the rigorous internal training and validation strategy supports the robustness of the findings and demonstrates that ML can detect subtle vascular signatures that remain inaccessible to conventional clinical assessment. 
Finally, although standardized imaging protocols were applied during carotid ultrasonography, systematic variability introduced by differences in ultrasound platforms, device configurations, or software implementations could not be addressed as the cohort used utilized a single machine for all examinations, which may have influenced the extracted imaging features.

\section{Conclusion}
Our findings demonstrate that the model learns biologically plausible and visually interpretable representations that capture vascular damage far beyond the information encoded in the hypertension label used for training. 
These representations exhibit robust associations across multiple domains: they correlate strongly with established clinical parameters and comorbidities, stratify long-term mortality more effectively than hypertension status itself, and provide substantial predictive value in Cox proportional hazards models.
Notably, the VD marker matches or outperforms traditional clinical benchmarks such as SCORE2, indicating that the model detects structural and functional vascular abnormalities that reflect genuine cardiovascular risk rather than merely reproducing hypertension status.

Importantly, this study shows that routine carotid ultrasound contains far more diagnostic and prognostic information than currently leveraged in clinical practice.
Despite being trained with weak and inherently noisy supervision, the model uncovers clinically relevant risk signals that rival—and in several scenarios exceed—the predictive accuracy of established risk models that require laboratory data and are restricted to specific age ranges.
The ability to extract such rich information directly from B-mode ultrasound underscores the substantial untapped potential of this widely available imaging modality.

These insights highlight the possible clinical impact of our approach.
A fully automated, video-based model capable of quantifying VD offers a scalable, non-invasive, and cost-effective means of cardiovascular risk assessment.
It could support early identification of high-risk individuals in settings where laboratory based models are difficult to implement, complement existing tools by capturing aspects of vascular pathology they overlook, and ultimately contribute to more personalized and equitable prevention strategies.
By enhancing the informational value of a standard examination already performed millions of times each year, this work points toward a future in which carotid ultrasound becomes a more powerful and accessible instrument for cardiovascular risk prediction.

\section{Methods}
\subsection{Gutenberg Health Study Dataset}
The Gutenberg Health Study (GHS) \cite{wild2012gutenberg} is a large-scale, prospective population-based cohort study initiated in April 2007 by University Medical Center Mainz.
With 15,010 participants in the initial baseline recruitment wave, and by now over unique 20,000 participants, it is among the largest population-based prospective cohort studies globally. 
The study focuses on the health status and disease progression within the Rhine-Main region (Germany), with primary emphasis on cardiovascular health. 
Its goal is to identify risk factors and causes of common diseases, contributing to preventive healthcare. 
As part of these comprehensive evaluations, each individual underwent a carotid sonography assessment, where multiple ultrasound videos from different perspectives were collected.
In addition to the ultrasound assessment, individuals were screened for anthropometric characteristics, traditional risk markers, comorbidities, laboratory parameters, and also future incidents. 
Additional details regarding the study design can be found in a separate publication \cite{wild2012gutenberg}.

\subsection{Cardiovascular Risk Factors and Endpoint Definitions}\label{sec:hyper_definition}
Hypertension is defined as the combination of systolic blood pressure above or equal to 140 mmHg, diastolic blood pressure above or equal to 90 mmHg, and/or the self-reported intake of anti-hypertensive medication during the last two weeks prior to inclusion in the study.
Comorbidities include atrial fibrillation, congestive heart failure, past myocardial infarction (MI), past stroke, coronary artery disease, and CVD.
Risk factors include dyslipidemia, diabetes type 2 and SCORE2-Germany (applicable to individuals without prior CVD or diabetes, aged 40–69 years).
The presence of risk factors is determined by a combination of self-reported information, inhouse biomarker measurements as well as medication intake.
Incidence data on cardiovascular events were assessed via structured follow-ups with subsequent validation of endpoints.
All-cause death was obtained via monthly checks with German registration offices.
Cardiac death was determined via quarterly review of death certificates using ICD-10 coding.

\subsection{Hypertension as Noisy Proxy}
Direct inference of absolute blood pressure from grayscale B-mode ultrasound is not feasible in the absence of Doppler imaging.
Accordingly, we conceptualize hypertension as a noisy proxy label for ML purposes: the model is trained to predict the label of hypertension indirectly, by identifying structural and functional vascular alterations that are associated with chronically elevated blood pressure.
Label noise arises as a consequence of the fact that not all individuals classified as hypertensive exhibit detectable vascular changes, while vascular damage may also result from other conditions. 
Factors such as treatment, disease duration, and phenomena like white-coat hypertension further contribute to variability. 
From a ML perspective, the model thus learns to identify structural and functional consequences of elevated blood pressure rather than the absolute blood pressure.

\subsection{Architecture Selection}
To balance computational efficiency and predictive performance, we evaluated three ML architectures prior to cross-validation: VideoMAE \cite{tong2022videomae}, ViViT \cite{arnab2021vivit}, and TimeSformer \cite{bertasius2021space}, all originally developed for general-purpose video classification. 
Each model was fine-tuned on the task of predicting hypertension from carotid ultrasound videos. 
VideoMAE achieved the highest validation F1-score ($0.733$), although performance across all architectures was comparable (Table \ref{tab:performance}).
Given its lower computational requirements, VideoMAE provides the most favorable trade-off between efficiency and accuracy and was therefore selected as the backbone for all subsequent analyses.

\subsection{Dataset split}
The dataset comprises 14,741 participants with at least one valid ultrasound video. 
Due to the low prevalence of certain outcomes, such as stroke (N = 176), the dataset was partitioned into 10 randomly drawn, non-overlapping test sets.
For each split, the remaining 90\% of the data were further divided into 80\% training and 20\% validation subsets.
All reported results are based on the aggregated performance across the 10 independent test sets, ensuring robust estimation of model generalization.

\begin{table}[!b]
    \centering
    \begin{tabular}{c|ccc}
        Model & $bACC_{val}\uparrow$ & $F_{1,val}\uparrow$ & $AUC_{ROC,val}\uparrow$ \\ \hline
        VideoMAE & 70.61\% & \textbf{0.733} & \textbf{0.773} \\ 
        ViViT & \textbf{70.88}\%  & 0.709 & 0.772 \\
        TimeSformer & 69.07\% & 0.724 & 0.763 \\
    \end{tabular}
    \caption{Individual-level accuracy and balanced accuracy on the validation set for different backbone architectures}
    \label{tab:performance}
\end{table}

\subsection{Model Platform \& Preprocessing}
All experiments were performed using the Huggingface Trainer Class Interface. 
The same is also used as a source for pretrained model weights, which are used for initialisation of the models before finetuning on GHS data.
The raw videos extracted from the DICOM files initially retained elements of the ultrasound device’s user interface, including the heartline displayed on-screen. 
The user interface and heartline were removed, and videos containing Doppler visualizations were excluded from the dataset.
These steps were implemented to avoid biases or misinterpretations.

From each processed video, multiple clips were uniformly sampled without overlap. 
The videos were normalized using the mean and standard deviation values estimated during the pre-training.
During finetuning, data augmentation techniques were applied, including random short-side scaling, random cropping to a resolution of $224 \times 224$, and random horizontal flipping. 
Weighted random sampling was employed to address class imbalances in the dataset. 
Training was conducted over ten epochs, with evaluations performed after each epoch. 
The final model was selected based on the best-performing evaluation results.

\subsection{Model interpretation}
To assess the relevance and clinical plausibility of the features learned by the model, we used two complementary interpretability approaches: occlusion-based feature attribution and counterfactual example generation. 
Both methods provide insight into the image regions and latent concepts that influence the model’s predictions.
We qualitatively compared the outputs of both techniques and additionally used attribution averaging and statistical analyses to quantify the highlighted features.

\subsubsection{Occlusion-Based Attribution}
Occlusion-based attribution produces maps indicating how strongly each pixel contributes to the model’s decision.
Using masked video occlusion, we identified regions that were consistently important across the entire sequence; spatio-temporal occlusion was used to examine how model attention shifts during the cardiac cycle.
Both methods were implemented with the Captum library \cite{captum}.

For masked occlusion analysis, attributions were computed for all videos and grouped by anatomical region (CCA, ICA, ECA), side (left, right), and assessment time point (baseline, 5-year, 10-year).
To obtain representative averages without losing all anatomical structure, we selected 50 videos per group that were most similar to the global dataset mean.
This approach reduces blurring due to anatomical variability while limiting selection bias.
From these selected videos, we computed mean, median, and top-5\% positive and negative attribution maps to summarize the most influential regions.

\subsubsection{Counterfactual Examples}
Counterfactual examples illustrate how an input must change for the model to switch its prediction.
They reveal which visual features are essential for distinguishing between low and high VD.
We generated counterfactuals using LD-ViCE \cite{varshney2025ld}, which produces minimally modified video representations that invert the model output.
This enables direct examination of structural changes that move the prediction across the decision boundary.

\subsection{Statistical Comparison}
We applied statistical models, to evaluate the alignment of the model’s extracted features with cardiovascular risk.
Therefore, we conducted statistical comparisons of hypertension (hypertensive vs. non-hypertensive) and the degree of VD (low vs. high VD). 
To assess the arterial health condition of each individual group, we analyse and compare various clinical parameters across the following categories: comorbidities, risk factors, the total plaque count, IMT (left and right) and arterial stiffness obtained from carotid sonography, and laboratory assessed Troponin I and NT-proBNP.

For each parameter, the distribution of values (represented by quartiles: 25\%, median, 75\%) or the prevalence of specific conditions is systematically compared across groups.

\subsection{Traditional Statistical Approaches}
To contextualize the model-derived VD marker within established cardiovascular risk assessment frameworks, we applied two conventional risk stratification methods and conducted survival analyses using Cox proportional hazards modelling.
These analyses were used to evaluate the independent and additive prognostic value of the imaging-based VD estimate.

\subsubsection{Baseline Risk Stratification Using SCORE2}
As a standard clinical benchmark, we computed SCORE2 \cite{esc2021score2}, a 10-year risk prediction model developed for European adults aged 40–69 years without prior cardiovascular disease or diabetes.
SCORE2 incorporates age, sex, smoking status, systolic blood pressure, and total and HDL cholesterol, along with region-specific calibration parameters.

For each eligible participant, the SCORE2 10-year risk estimate was calculated and used as a reference against which to compare the predictive performance of our VD marker.
All analyses involving SCORE2 were restricted to participants with complete and valid SCORE2 parameters, which resulted in a reduced sample size and corresponding reduction in observed events.

\subsubsection{Kaplan-Meier Estimates}
Kaplan–Meier survival curves were used to compare event-free survival across different risk strata.
We stratified participants into high- and low-risk groups based on the model’s VD prediction, and compared these curves with stratifications derived from (i) the hypertension label and (ii) SCORE2 as a state-of-the-art clinical risk model.

For SCORE2, we used the recommended threshold of 8\cite{esc2021score2} to distinguish between high and low estimated 10-year cardiovascular risk.
Survival curves were generated for outcomes including cardiac death (5- and 10-year horizons), stroke, MI and all-cause mortality.

\subsubsection{Cox Proportional Hazard Modelling}
Cox proportional hazards regression \cite{cox1972regression} was used to quantify the association between the VD marker and time-to-event outcomes. 
The model estimates hazard ratios for covariates under the assumption of proportional hazards, and its performance is evaluated using the C-index, which measures the correctness of the predicted risk ordering.

For the VD marker, we used the model’s continuous prediction confidence rather than its binary class label.
The confidence scores were linearly scaled to the interval $[0,1]$ and normalized to unit standard deviation; the same normalization procedure was applied to all covariates.
Statistical significance of C-index differences between models was assessed using established significance tests \cite{kang2015comparing}.

\begin{credits}
\subsubsection{Ethics and Consent to Participate}
The data analysed in this study were obtained from the Gutenberg Health Study (GHS), an interdisciplinary population-based cohort study. 
The GHS was approved and is continuously monitored by the Ethics Committee of the Rhineland-Palatinate Medical Association, the data protection officer of the Mainz University Medical Center, and the Rhineland-Palatinate State Commissioner for Data Protection and Information Security.

All participants provided written informed consent before participating in the study. 
The specific analysis presented here was approved by the GHS Steering Committee (Proposal: GHS2025\_EX001). 
All research activities were conducted in accordance with the guidelines of the 1964 Declaration of Helsinki and its later amendments, as well as the principles of Good Scientific Practice and Good Epidemiological Practice.

\subsubsection{Data Availability Statement}
The data that support the findings of this study (including raw carotid ultrasound videos and linked clinical records) are not publicly available due to legal and ethical restrictions regarding the privacy of research participants. 
However, the de-identified datasets are accessible for scientific research purposes. 
Requests for data access should be submitted via the official study website [https://www.unimedizin-mainz.de/ghs/en/informationen-for-scientists/access-to-study-data-and-biomaterial.html]. 
Access is subject to approval by the steering committee, the execution of a formal data use agreement, and, where applicable, institutional review board approval.

\subsubsection{Author Contributions Statement}
C.B. conceived and designed the study, developed the core machine learning framework, performed data preprocessing and primary experiments, conducted the statistical analysis, and wrote the original draft of the manuscript. 
A.R. provided clinical expertise, supervised the clinical phenotyping of the cohort, and wrote specific sections of the manuscript regarding the clinical aspects of hypertension and cohort characteristics. 
P.V. designed and implemented the computational pipeline for generating the counterfactual examples and contributed to the explainable AI (XAI) methodology. 
V.T., K.G., J.T., P.C., A.S., D.T., K.K. and P.W. contributed to the acquisition of clinical data and provided guidance on the medical evaluation of vascular damage. 
S.A. and A.D. provided technical guidance on the machine learning architecture and data validation strategies. 
All authors contributed to the interpretation of the results, performed critical proof-reading and editing of the manuscript, and approved the final version for submission.

\subsubsection{\discintname}
The authors have no competing interests to declare that are relevant to the content of this article.

\subsubsection{Declaration of generative AI and AI-assisted technologies in the writing process}
During the preparation of this manuscript, the authors used Gemini (Google) in order to improve the readability and language of the text.
After using this tool, the authors reviewed and edited the content as needed and take full responsibility for the content of the published article.

\subsubsection{\ackname} 
This work was part of the cluster for atherothrombosis and individualized medicine (curATime), funded by the German Federal Ministry of Research, Technology and Space (03ZU1202KA).

\begin{figure*}[h]
 \centering
    \includegraphics[width=.45\linewidth,trim=0cm 0cm 0cm 0cm]{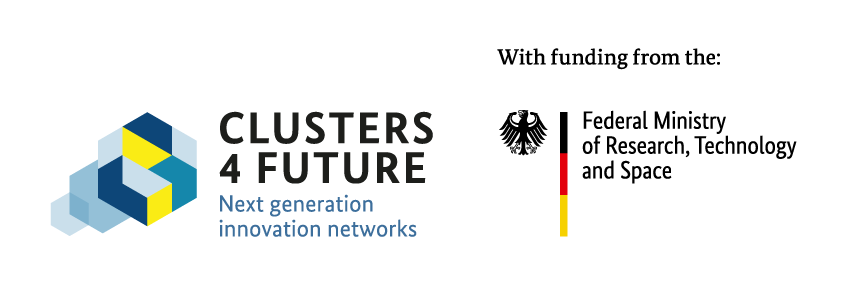}
\end{figure*}
\end{credits}


\bibliographystyle{splncs04} 
\bibliography{refs} % Entries are in the refs.bib file

\end{document}